\documentclass[10pt,twocolumn,letterpaper]{article}

\usepackage[pagenumbers]{cvpr} 
\usepackage{placeins}
\usepackage{graphicx}

\usepackage{url}  
\usepackage{xspace}
\usepackage{multirow}

\newcommand{\snp}{{\sc Snp}\xspace}
\newcommand{\siren}{{\sc Siren}\xspace}

\newcommand{\strainer}{{\sc Strainer}\xspace}

\definecolor{cvprblue}{rgb}{0.21,0.49,0.74}
\usepackage{float}
\usepackage[pagebackref,breaklinks,colorlinks,allcolors=cvprblue]{hyperref}

\newcommand{\snpinr}{{\sc Snp}\xspace}

\newcommand{\snpB}{{\sc \textbf{Snp}}\xspace}
\newcommand{\papertitle}{The Surprising Effectiveness of Noise Pretraining \\ for Implicit Neural Representations}

\newcommand{\stnoise}{{structured noise}\xspace}

\newcommand{\celeba}{CelebA-HQ\xspace}

\newcommand{\ipc}{IPC}

\newcommand{\nceleba}{550\ }
\newcommand{\nafhq}{368\ }
\newcommand{\nmri}{144 \ }

\newcommand{\stt}{structured\xspace}
\newcommand{\ustt}{unstructured\xspace}

\newcommand{\afhq}{AFHQ\xspace}
\newcommand{\mri}{OASIS-MRI\xspace}

\def\paperID{XXXXX} 
\def\confName{CVPR}
\def\confYear{2026}

\title{\papertitle}

\author{
Kushal Vyas$^{1}$ \quad Alper Kayabasi$^{2}$ \quad Daniel Kim$^{1}$ \quad Vishwanath Saragadam$^{2}$ \\  Ashok Veeraraghavan$^{1}$ \quad Guha Balakrishnan$^{1}$\\
{$^1$ Rice University \quad $^2$ University of California, Riverside}\\ 
{\tt\small \{kvyas,dk77,vashok,guha\}@rice.edu} \quad {\tt\small \{akaya003, vishwans\}@ucr.edu} 
}

\begin{document}
\twocolumn[{%
\renewcommand\twocolumn[1][]{#1}%
\maketitle
\begin{center}
    \centering
    \captionsetup{type=figure}
    \includegraphics[width=\textwidth]{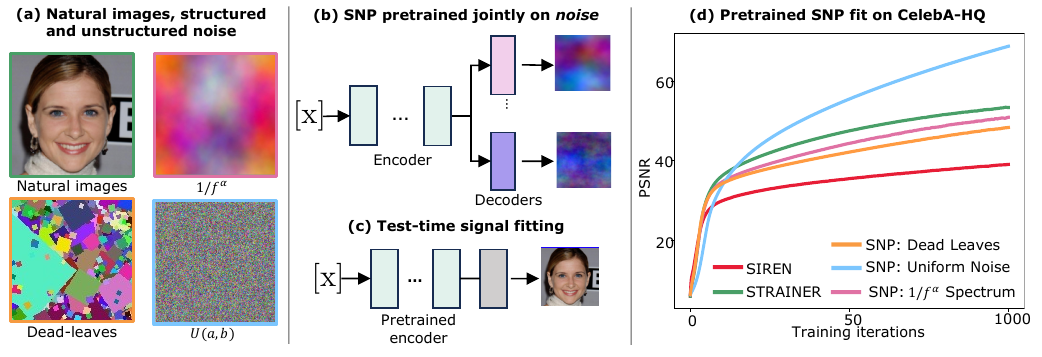}
    \captionof{figure}{\textbf{In this study, we explore the effectiveness of initializing INR parameters with signals from noise classes with different properties.}
    (a) Samples of a few of the signal types considered in this study: natural images, noise with the classic ``$1/|f^\alpha|$'' spectral signature of natural images, Dead Leaves noise, and Uniform noise. (b) We propose \snp, a simple but powerful data-driven INR parameter initialization method using noise samples leveraging the \strainer~\cite{vyas_strainer} initialization framework. We pretrain an INR encoder with signal-specific decoder heads on noise samples from each type. (c) At test time, the pretrained encoder with a randomized decoder is fine-tuned on a new (real) signal. (d) Surprising test-time signal fitting results on \celeba: \snp initialization using Uniform noise performs dramatically better on signal fitting compared to all other methods.}
    \label{fig:teaser}
\end{center}
}]

\begin{abstract}
The approximation and convergence properties of implicit neural representations (INRs) are known to be highly sensitive to parameter initialization strategies. While several data-driven initialization methods demonstrate significant improvements over standard random sampling, the reasons for their success -- specifically, whether they encode classical statistical signal priors or more complex features -- remain poorly understood. In this study, we explore this phenomenon through a series of experimental analyses leveraging noise pretraining. We pretrain INRs on diverse noise classes (e.g., Gaussian, Dead Leaves, Spectral) and measure their ability to both fit unseen signals and encode priors for an inverse imaging task (denoising). Our analyses on image and video data reveal a surprising finding: simply pretraining on unstructured noise (Uniform, Gaussian) dramatically improves signal fitting capacity compared to all other baselines. However, unstructured noise also yields poor deep image priors for denoising. In contrast, we also find that noise with the classic $1/|f^\alpha|$ spectral structure of natural images achieves an excellent balance of signal fitting and inverse imaging capabilities, performing on par with the best data-driven initialization methods. This finding enables more efficient INR training in applications lacking sufficient prior domain-specific data. 
For more details, \href{https://kushalvyas.github.io/noisepretraining.html}{visit the project page!}
\end{abstract}    
\section{Introduction}
\label{sec:intro}
Implicit neural representations (INRs) offer high-fidelity, continuous signal representations that have proven useful for various computer vision and graphics applications such as compression~\cite{dupont2022coin++,kayabasi2025bias,saragadam2022miner}, inverse imaging~\cite{saragadam2023wire,sitzmann2020implicit}, and neural rendering~\cite{mildenhall2020nerf}. Typically implemented as multilayer perceptrons (MLPs) mapping spatial coordinates to signal values, INRs are traditionally initialized with weights drawn from standard Gaussian or Uniform distributions. Yet, despite their straightforward architecture, INRs are highly sensitive to activation functions and initialization strategies, both of which strongly dictate convergence rates. In particular, recent studies show that data-driven approaches to parameter initialization relying on prior training signals, such as meta-learning, hyper-networks, and transfer learning, significantly outperform standard random sampling strategies in terms of signal fitting performance and encoding ``deep'' priors for solving inverse imaging tasks such as denoising and super-resolution~\cite{functa,sitzmann2020implicit,vyas2025fit,vyas_strainer, ipc}. 

In this study, we investigate \emph{why initializing INR parameters using prior data is so effective.} In particular, based on findings that INRs have low-frequency spectral biases~\cite{xie2023diner,rahaman2019spectral,tancik2020fourfeat} and that initialization succeeds even using out-of-distribution training signals~\cite{vyas_strainer}, we ask: does the benefit of data-driven initialization stem from encoding dataset-specific mid-to-high-level features, or from simpler universal natural image statistics—such as location-independent strong edges and a $1/|f^\alpha|$ spectral structure~\cite{van1996modelling}, or even something else? Furthermore, if simple properties are indeed most responsible for the success of data-driven initialization strategies, \emph{is it possible to develop equivalently powerful that do not require prior real data}? Doing so would enable one to train INRs efficiently in data-poor applications often found in scientific imaging scenarios. 

Directly investigating the properties of initialized INRs to answer these questions is challenging due to their high-dimensional intermediate feature spaces. Instead, we adopt an empirical approach inspired by recent work demonstrating the efficacy of pretraining visual recognition networks on noise variants, or pseudo-natural images generated by imposing well-studied natural image statistics on uniform noise~\cite{baradad2021learning,baradad2022procedural,KataokaIJCV2022,Kataoka_2022_CVPR,yu2024learning}. Specifically, we initialize INR parameters using signals from distinct noise classes (e.g., Gaussian, pink noise, dead leaves) and measure the resulting convergence properties when fitting real test signals. Because each noise class is defined by a precise, succinct set of properties, we hypothesize that this analysis can reveal which fundamental, low-level signal characteristics, if any, drive the performance gains observed in data-driven INR initialization.

We conducted a series of experimental analyses on image and video data, focusing on a simple yet effective data-driven initialization approach called \strainer~\cite{vyas_strainer}. This method divides a multilayer perceptron (MLP) into a shared \text{encoder} and $N$ individual \textit{decoder} heads, training the network on a set of $N$ real signals (in our experiments, as few as $N \approx 10$ signals). 
At test time, the pretrained encoder is combined with a randomly initialized decoder to fit novel signals with remarkable efficiency. We name our variant of \strainer, which is instead pretrained on noise samples, as \snp (for \emph{\strainer Noise Pretraining}).

Using image datasets, we first present the surprising finding that \snp pretrained on \emph{highly unstructured noise} (e.g., Uniform and Gaussian) achieves remarkably higher signal fitting performance than when pretrained on structured noise or real images (e.g., reaching $\approx 80$ dB PSNR compared to $\approx 50$ dB at convergence). Similarly, \snp results for video fitting demonstrate that unstructured noise pretraining considerably improves the reconstruction quality of a standard video INR model (ResFields~\cite{resfield}). Crucially, however, we find that this unstructured noise yields INRs with poor denoising capabilities compared to both \strainer and \siren, indicating a lack of sufficient deep image priors. This highlights a fundamental tradeoff between signal fitting capacity and prior strength when selecting parameter initialization strategies. In contrast, pretraining on noise with a $1/|f^\alpha|$ spectral structure offers an excellent balance: it achieves 
nearly equivalent signal fitting and denoising performance to \strainer.
Importantly, because this noise is easily generated, it eliminates the need for real data samples to effectively initialize INRs. These findings, which generalize across natural image and video domains, provide fundamental insights into INR behavior that can guide future research and engineering practices.

\section{Related Work} 
\label{sec:background}
INRs have been widely studied and used for signal representation tasks for audio~\cite{lanzendorfer2023siamese,sitzmann2020implicit}, images~\cite{tancik2020fourfeat,sitzmann2020implicit,functa,ngp,xie2023diner}, video~\cite{ghorbel2024nerv,actinr,rangarajan2025siedd} and multi-dimensional data~\cite{mildenhall2020nerf,ngp,martinbrualla2020nerfw}. Furthermore, since INRs can be easily incorporated with underlying physical models, they have had natural applications for inverse imaging tasks in computer vision such as novel view synthesis~\cite{mildenhall2020nerf,ngp} and sparse-view CT reconstruction in medical imaging~\cite{naf, Zang2021IntraTomo}. 

\textbf{Initialization of INRs.} INRs are extremely sensitive to parameter initialization, which dictates their ability to converge to a target signal. Unlike standard deep networks that can be robustly initialized using He~\cite{he_init} or Xavier~\cite{pmlr-v9-glorot10a} methods, INRs require specialized tuning. Sitzmann et al.~\cite{sitzmann2020implicit} proposed a principled initialization scheme for \siren that remains the most widely used baseline for INRs. Building on this, methods such as Functa~\cite{functa}, Spatial Functa~\cite{scalingfuncta}, LearnIt~\cite{tancik2021learned}, and TransINR~\cite{transinr} offer scalable ways to meta-learn initializations that generalize to unseen signals within a few gradient steps at test time. Additionally, the Implicit Pattern Composers (IPC) framework~\cite{ipc} introduces an INR design where only the second-layer weights are modulated to represent a signal, utilizing a transformer meta-learner over a distribution of natural images to achieve high-quality representations. Finally, \strainer~\cite{vyas_strainer} demonstrates that an INR with shared initial \textit{encoder} layers and deeper, signal-specific \textit{decoder} heads learns highly transferable features. Surprisingly, this work also shows that encoder features trained on one domain (e.g., human faces) serve as excellent initializations for fitting out-of-distribution signals such as animal faces and MRI scans. This unexpected result suggests that INR parameter initialization plays a much broader role in capturing universal features that generalize across natural images.

\textbf{Inductive biases of INRs.} INRs equipped with ReLU activations are notoriously prone to spectral biases, favoring low frequencies and struggling to capture high-frequency detail. SIREN~\cite{sitzmann2020implicit} mitigates this issue by introducing sinusoidal activations, which substantially improve the reconstruction of fine structures. In practice, however, SIREN often fails to utilize the entire available frequency spectrum and requires a carefully designed initialization. FINER~\cite{liu2023finer} extends this approach using variable-frequency activations controlled by network biases. Concurrently, work on Gaussian activations reveals that the key mechanism for improved performance is not periodicity itself, but rather the modulation of second-order derivatives governing local smoothness. Nevertheless, both of these designs demand delicate initialization schemes and extensive hyperparameter tuning. WIRE~\cite{saragadam2023wire} proposes wavelet-based activations, leveraging the well-known effectiveness of wavelets as a basis for natural images and capturing denoising priors. Despite these various advances, all these methods remain tightly coupled to strict initialization rules, sensitive hyperparameters, and long training schedules.

\textbf{Training Neural Networks on Noise.} Deep neural networks are conventionally trained on real or semi-real signals. However, a line of work in computer vision~\cite{baradad2021learning,baradad2022procedural,KataokaIJCV2022,Kataoka_2022_CVPR,yu2024learning} demonstrate how structured noise -- pseudo-natural images obtained by imposing natural image statistics on random uniform noise -- may be used to pretrain image and video classification networks with useful features applicable for inference on real signals. These results indicate that much of the crucial visual cues leveraged by deep neural networks may in fact be low-level statistics and patterns of images rather than high-level features. Baradad et.al~\cite{baradad2021learning,baradad2022procedural} introduced the \emph{Looking at Noise} dataset~\cite{baradad2021learning} which contains a collection of procedurally generated \stnoise images obtained by imposing natural image statistics on random uniform noise. We use that dataset in this study.

\section{Experimental Analysis} \label{sec:experiments_results}
\begin{figure*}[!t]
    \includegraphics[width=\linewidth]{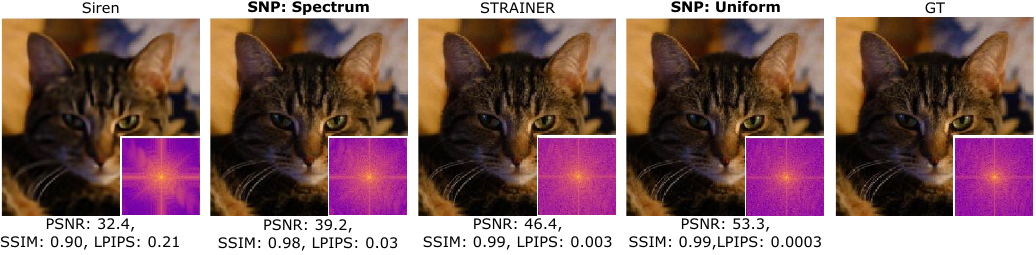}
    \caption{\textbf{Early reconstruction quality at $T=200$ steps.} \strainer, \snp (Spectrum and Uniform) learn high quality reconstructions in $200$ steps (PSNR$\approx40db+$), while the \siren output remains blurry. Power spectrum insets show that: \snp: Uniform matches closely with the ground truth, \strainer has a noisy high-frequency coverage, and \snp: Spectrum recovers most high frequencies while \siren lacks critical frequencies.}
    \label{fig:noise_recon_200}
\end{figure*}

We conducted a series of experiments on image and video data to explore the effectiveness of initializing INR parameters using various noise classes. We selected the \strainer~\cite{vyas_strainer} data-driven initialization method (see Fig.~\ref{fig:teaser}b) because it demonstrates superior performance over competing baselines, such as TransINR~\cite{transinr}, IPC~\cite{ipc}, and meta-learned INR initializations~\cite{tancik2021learned}. Furthermore, we empirically found that several of these meta-learning and hypernetwork baselines depend on a certain level of structure in the input signals, and are therefore not amenable to training over noise. 

We name the \strainer method pretrained on noise signals \snpinr (Fig.~\ref{fig:teaser}b). In all experiments, we implemented \snp as a $L=6$ layer MLP with sine activations~\cite{sitzmann2020implicit}($\omega=30$), with $L-1$ \emph{encoder} layers shared across signals and separate signal-specific final \emph{decoder} layers. During pretraining, we jointly fit \snp to $N=10$ signals for $T=5000$ iterations, identical to the procedure used in \strainer. During testing (fitting an unseen signal), we retained the pretrained weights of the \emph{encoder} layers, randomly initialized a single \emph{decoder} head, and fit both the encoder and decoder on the signal for $T=2000$ steps with an $L_2$ loss function. 
We use the Alpine INR framework~\cite{vyasalpine} for pretraining and test-time signal fitting for \snp INRs.
For video fitting, we additionally share MLP weights across frames, with learnable time-conditioned (frame-specific) residual updates, as used in ResFields~\cite{resfield}. We used the Adam~\cite{kingma2017adam} optimizer with a learning rate of $10^{-4}$. Please refer to Sec.~\ref{supl:train_fit_derivation} in Supplementary for further details on \strainer and \snp's pretraining and fitting process and Sections~\ref{sec:supl_finer}, \ref{sec:supl_addnexplayers} in Supplementary for additional quantitative results on different activations and network layer depths. 

\subsection{Noise Signals}
We considered a variety of noise types, both \stt and \ustt, visually illustrated in Sec~\ref{supl:supl_viewdataset} of Supplementary. For \ustt noise, we used Uniform and Gaussian. For Uniform, we sampled pixel intensities from $\mathcal{U}(0,1)$ and for Gaussian, we sampled pixel intensities from $\mathcal{N}(0.5, 0.2)$. For \stt noise, we used a subset of the \emph{Looking at Noise} dataset~\cite{baradad2021learning}, which consists of various types of \stnoise generated by carefully imposing natural image properties on Uniform noise. Specifically, we used the \emph{dead-leaves (squares, oriented, mixed, textures)} and statistical \emph{(spectrum, spectrum + color, wavelet marginal model or wmm, spectrum + wmm + color)} classes. These broad groups of noises cover several key natural image properties in the spatial and frequency domains such as strong edges and redundant patch statistics (seen in dead-leaves), and non-Gaussian spectral distributions (seen in statistical models). 

\subsection{Evaluation Datasets, Baselines, and Metrics}
We evaluated noise-pretrained \snp{}s on image and video representation and denoising tasks. For images, we used \nceleba \celeba~\cite{celebahq} faces, \nafhq (AFHQ)~\cite{afhq} images, and \nmri OASIS-MRI~\cite{oasis1,oasis2} images. For videos, we curated 10 video sequences with diverse motions (e.g., crowd, traffic, spinning coins), each approximately 10 seconds long and captured at 25 FPS, from the Pexels Video Repository~\cite{pexels}. We resized all images to $178 \times 178$, maintaining their aspect ratios, and center-cropped all video frames to $512 \times 512$. For images, we evaluated \snp against standard (randomly initialized) \siren~\cite{sitzmann2020implicit} INRs, and INRs initialized with several parameter initialization methods: \strainer~\cite{stanfordcars}, meta learning~\cite{tancik2021learned}, TransINR~\cite{transinr}, and Implicit Pattern Composers~\cite{ipc}. We evaluated signal reconstruction results using PSNR (dB), SSIM, and LPIPS. 

\subsection{Image and Video Fitting Results}
We report mean reconstruction PSNR values for image fitting (at $T=2000$ iterations) on all image datasets in Table~\ref{tab:metrics_jumbo}, and PSNR as a function of iteration when fitting \celeba images in Fig.~\ref{fig:results_celeba_psnr_plot} (curves for \afhq and \mri are in Supplementary). Because \snp INRs and baselines converge to PSNRs of over 50dB in our image experiments, we omit SSIM and LPIPS values in the main text because they provide limited additional insight. However, we provide values for all metrics in Supplementary (see Sec.~\ref{more_results_image_fitting}).

\begin{figure}[t!]
    \centering
    \includegraphics[width=\linewidth]{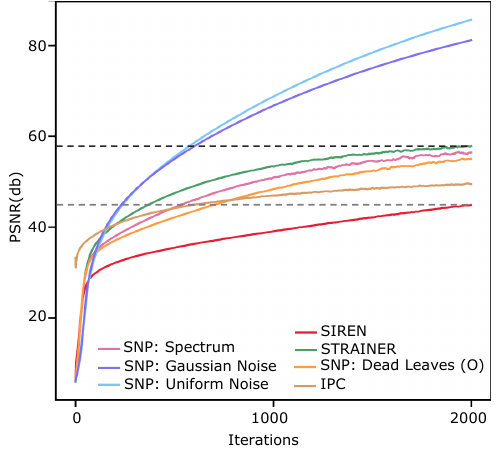}
    \caption{\textbf{Signal fitting results on \celeba images.} \snp initialized using various noises demonstrate strong signal representation ability and convergence compared to widely known baseline methods such as \siren~\cite{sitzmann2020implicit}, \ipc~\cite{ipc}, and \strainer~\cite{vyas_strainer}. Interestingly, \snp pretrained with Gaussian and Uniform noises performed significantly better than all other approaches, converging rapidly to perfect fitting.}
    \label{fig:results_celeba_psnr_plot}
\end{figure}

\begin{table}[ht!]
\centering 
\setlength{\tabcolsep}{1.8pt}
\caption{\textbf{Image fitting results.} We present PSNR values (in dB) for all test methods at $T=2000$ iterations, from vanilla randomized initialization (SIREN) to data-driven initialization methods (\strainer, Metalearned~\cite{tancik2020fourfeat}, TransINR~\cite{transinr}, IPC~\cite{ipc}) to noise pretraining with \strainer (\snp). We provide the pretraining dataset name in parentheses. \snp with unstructured noises (Uniform, Gaussian) perform surprisingly best, with remarkably higher PSNR values than all other methods.}

    \scalebox{0.9}{
    \begin{tabular}{@{}l@{\hspace{-6pt}}ccc@{}} 
    \hline
    \textbf{Method} & \textbf{\celeba} & \textbf{\afhq} & \textbf{\mri} \\
    
    \hline
        \siren~\cite{sitzmann2020implicit} & 44.9 & 45.1 & 53.03 \\
        \strainer~\cite{vyas_strainer} & 57.8 & 58.0 & 62.80 \\
        Metalearned~\cite{tancik2021learned} & 53.1 & 53.3 & 67.0 \\
        TransINR~\cite{transinr} (\celeba) & 51.9 & 49.0 & 55.5 \\
        IPC~\cite{ipc} (\celeba) & 49.7 & 47.2 & 51.4 \\
        
    \hline
        
        \snp: Dead Leaves oriented & 55.0 & 55.4 & 59.0 \\
        \snp: Dead Leaves textures & 55.0 & 55.3 & 58.9 \\
    
        \snp: Spectrum & 56.4 & 56.2 & 60.0 \\
    \hline
        \textbf{\snp: Uniform} & \textbf{85.7} & \textbf{79.9} & \textbf{79.3} \\
        \snp: Gaussian & 80.0 & 77.0 & 79.1 \\
    \hline
    \end{tabular}}
    \label{tab:metrics_jumbo}
\end{table}

\begin{figure*}[!t]
    \includegraphics[width=\linewidth]{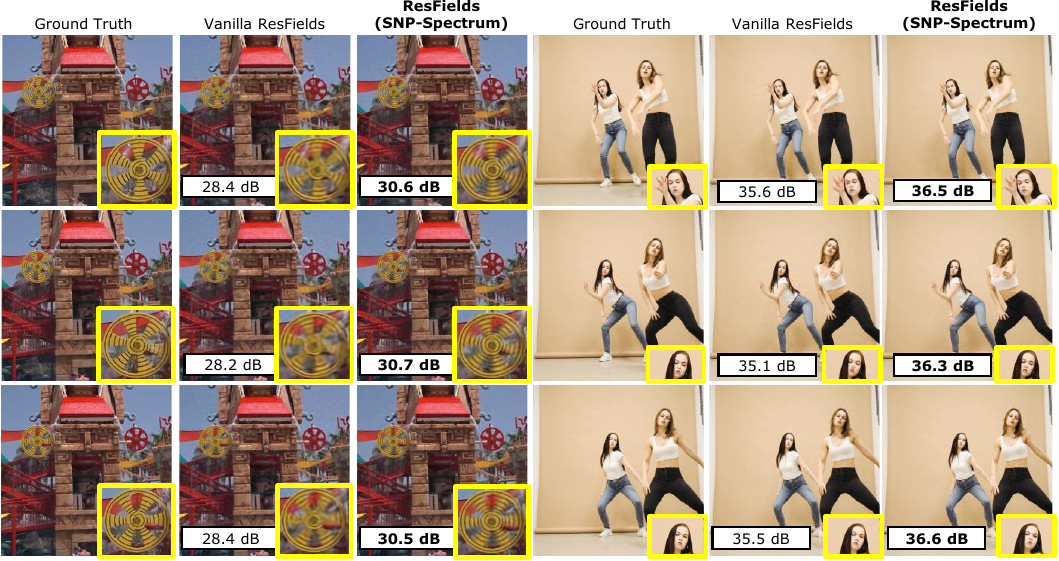}
    \caption{\textbf{Video fitting results: \snp improves fitting performance.} We show frames from representation of two example videos with randomly initialized (vanilla) ResFields~\cite{resfield}, and \snp: Spectrum-initialized Resfields. In both examples, \snp-initialization results in a higher accuracy with clearer visible details. This implies \snp advantages also expand beyond images to videos.}
    \label{fig:vid_fitting_samples}
\end{figure*}

\begin{figure*}[!t]
    \includegraphics[width=\linewidth]{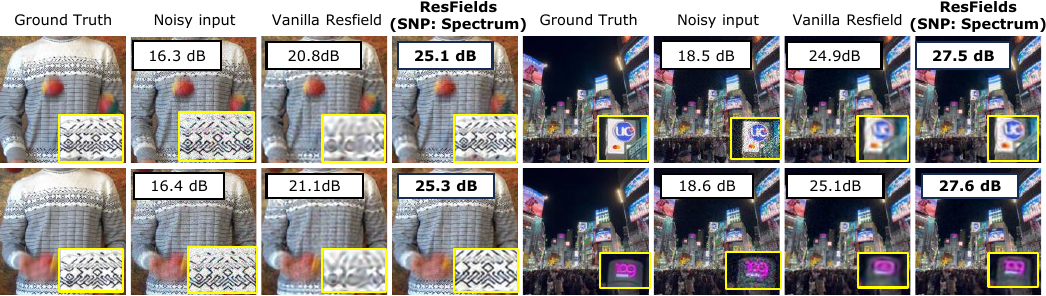}
    \caption{\textbf{Video denoising results.} The figure above shows denoised frames of sample videos with randomly initialized ResFields~\cite{resfield}, and \snp: Spectrum-initialized ResFields. In both examples, \snp: Spectrum outperforms by more than 2dB, with dramatically visual improvement, underscoring the effectiveness of \snp trained INRs for solving inverse problems.}
    \label{fig:vid_denoising_samples}
\end{figure*}

We observe the following general results for all image datasets. First, we see the striking result that \snp initialized with the \ustt noises (Uniform and Gaussian) achieve rapid convergence to high PSNRs ($\approx80db+$) far faster than all other methods, including vanilla \siren, all data-driven methods (\strainer, meta-learned~\cite{tancik2021learned}, TransINR~\cite{transinr}, IPC~\cite{ipc}), and other noise types. \strainer achieves excellent performance as expected, outperforming all other methods (other than for \mri) except unstructured noises. Of the structured noise initializations, \snp: Spectrum (noise exhibiting natural image spectral structure) performs best, almost matching \strainer. Fig.~\ref{fig:noise_recon_200} presents reconstructions for one image for several methods at an early fitting stage of $T=200$ along with their power spectra, showing that the different approaches have markedly different frequency characteristics.  

\begin{table}[t]
\centering
\caption{\textbf{Video fitting and denoising results.} We present mean PSNR values (in dB) for various methods at 100k iterations over all 10 videos from the Pexel video set. We use the ResFields~\cite{resfield} video INR framework, both with standard random parameter initialization and with \snp initialization for its per-frame weight parameters. \snp: Spectrum performs best for both fitting and denoising tasks. See Fig.~\ref{fig:vid_fitting_samples} for sample video fitting results.}
\label{tab:video_results}
\resizebox{\linewidth}{!}{%
\begin{tabular}{llc}
\hline
 \textbf{Task} & \textbf{Method}  & \textbf{Mean PSNR}\\
\hline
\multirow{3}{*}{Fitting}
& Vanilla ResFields~\cite{resfield}  & 29.5\\
& ResFields + \snp (Spectrum)  & 31.1\\
& ResFields + \snp (Uniform) & 30.9\\
\hline
\multirow{3}{*}{Denoising}
& Vanilla ResFields~\cite{resfield}    & 27.0\\
& ResFields + \snp(Spectrum)   & 28.0\\
& ResFields + \snp (Uniform) &  27.6\\
\hline
\end{tabular}%
}
\label{tab:video_results}
\end{table}

Table~\ref{tab:video_results} reports reconstruction PSNR (in dB) for the task of video fitting averaged across 10 selected sequences from the Pexels dataset (see Supplementary Table~\ref{tab:supl_video_detailed_results} for numerical results for individual video fitting and Fig.~\ref{fig:individual_fitting} for convergence plots). 
\snp-initialized ResFields models perform better than vanilla randomly initialized ResFields~\cite{resfield}, preserving delicate and fine structures with higher fidelity as shown in Figure~\ref{fig:vid_fitting_samples}, and converge to high quality video representation in significantly fewer number of steps (see reconstruction quality v/s iterations plot in Fig.~\ref{fig:vid_fitting_curves} in Supplementary).
Interestingly, unlike the case of image fitting, we find \snp: Spectrum performs slightly better than \snp: Uniform. 

\subsection{Image and Video Denoising Results}
Signal fitting and representation prioritizes the capacity of a given functional representation, but does not consider the \emph{inherent priors} of the representation to solve inverse imaging tasks. In this section, we analyze the performance of \snp models on denoising, a fundamental inverse imaging task. We present image denoising results in Table~\ref{tab:denoising_quantitative}. \siren performs best followed closely by \strainer and \snp: Spectrum, while \snp with unstructured noises (Uniform, Gaussian) perform worst. We also show in Fig.~\ref{fig:inv_tradeoff} that \snp: Spectrum and \strainer (pretrained on \celeba as well as ImageNet~\cite{imagenet}) which have more structured priors achieve better performance as early as 50 iterations whereas \snp (Uniform, Gaussian) which lack any structured prior and demonstrate superior signal representation perform poorly at denoising. We present video denoising quantitative results in Table~\ref{tab:video_results} and qualitative visual results in Fig.~\ref{fig:vid_denoising_samples}. ResFields initialized using \snp: Spectrum  also performs the best, followed closely by both ResFields + \snp: Uniform and vanilla ResField.

\begin{table}[t!]
\centering
\caption{\textbf{Denoising results for \celeba images at $T=2000$ iterations.} \strainer and \snp: Spectrum achieve similar denoising performance as \siren, but $2\times$ faster. \snp pretrained on unstructured noises (Uniform, Gaussian) perform poorly.}
\label{tab:denoising_quantitative}

\begin{tabular}{|c|c|c|c}
\hline
\multirow{2}{*}{\textbf{Method}} 
    & \multicolumn{2}{c|}{\textbf{Early Stopping}} \\
\cline{2-3}
 & PSNR & Iteration  \\
\hline
\siren~\cite{sitzmann2020implicit} & 28.3 & 139 \\
\strainer~\cite{vyas_strainer}     & 27.3 & 70 \\
\snp: Spectrum                     & 27.6 & 78 \\
\snp: Uniform                      & 23.0 & 73 \\
\snp: Gaussian                      & 23.8 & 70 \\
\hline
\end{tabular}
\end{table}

\subsection{Mechanistic Interpretability of \snp INRs} 
\label{sec:interpretation_analysis}
To understand why \snp works, we evaluate neural tangent kernels (NTKs) and local complexities of INRs initialized under different settings, and visualize their loss surfaces.

\begin{figure}[t!]
    \centering
    \includegraphics[width=\linewidth]{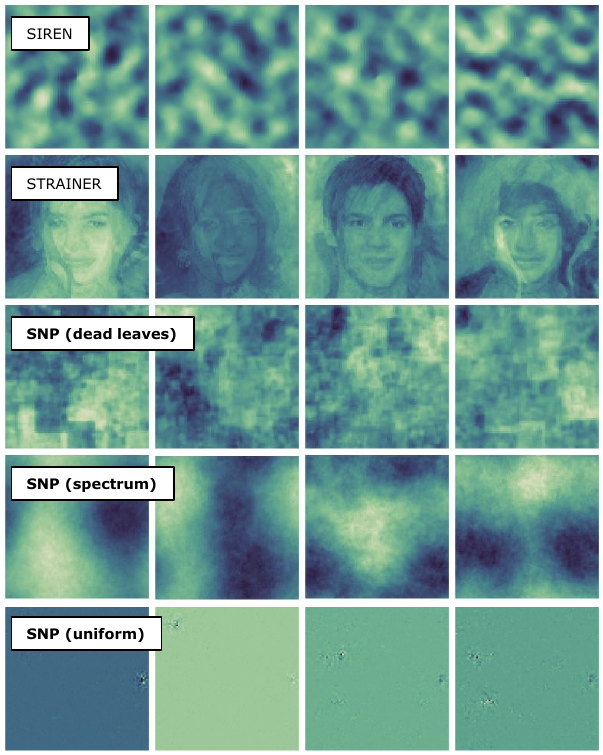}
    \caption{\textbf{NTK eigenvectors of INRs using different initialization schemes.} NTK eigenvectors characterize the function space spanned by a network in the vicinity of its initialization, and can help us depict how pretraining encodes priors in an INR's function space to influence its ability to fit signals. We visualize the top 4 NTK eigenvectors. Standard \siren initialization yields noticeably blurred and random eigenvectors. \strainer pretrained on \celeba exhibits face-like structures in its eigenvectors. \snp: Dead Leaves and \snp: Spectrum exhibits smooth regions akin to its structured noise images. Finally, \snp: Uniform pretrained on Uniform noise has essentially no noticeable structural features.} 
    \label{fig:ntk_eigvecs}
\end{figure}

\paragraph{NTK analysis.} NTK theory characterizes the evolution of a network’s outputs near initialization. Under a first-order linearization, network dynamics follow an autonomous linear ODE: $\frac{d f(t)}{dt} = -K\big(f(t)-y\big)$, where the NTK $K$ defines directions along which the network corrects its error, with convergence rates set by the corresponding eigenvalues~\cite{jacot2018neural,structured_dictionary}. Building on this view, we can interpret INRs as dictionaries whose bases are the NTK eigenvectors, where the fraction of a target signal’s energy captured by the top eigenmodes quantifies its learnability~\cite{structured_dictionary}.

\begin{figure}[t!]
    \centering
    \includegraphics[width=\linewidth]{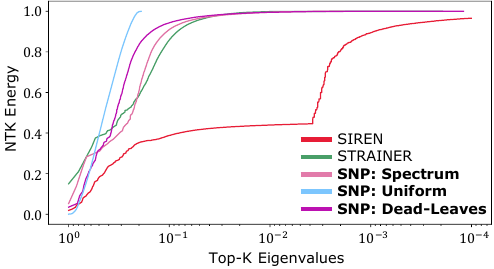}
    \caption{\textbf{Energy distribution across eigenmodes}. The $x$-axis denotes the eigenvalue percentile threshold; at a given point on the $x$-axis, we consider only eigenvalues above that percentile. The y-axis reports the fraction of the target-images energy captured by the eigenvectors associated with those eigenvalues. In other words, the curve summarizes how much cumulative energy is explained as we progressively include eigenvectors corresponding to eigenvalues in descending order. }
    \label{fig:ntk_curves}
\end{figure}

Figure~\ref{fig:ntk_curves} shows the energy captured by the top-$K$ eigenvalues of NTK for various representations. \snp: Uniform captures the NTK energy with the fewest eigenvalues, consistent with results from the previous experiments showing that \snp: Uniform has the fastest training times. Second, \strainer and \snp outperform randomly initialized \siren. Finally, \snp: Spectrum and \snp: Dead Leaves both converge comparably to \strainer, once again confirming the power of training INRs with noise. Figure~\ref{fig:ntk_eigvecs} further illustrates that \snp's atoms capture sharp, high-frequency discontinuities when trained on Dead Leaves textures, and span complex, non-convex structures when trained on stationary spectral noise. Overall, these characteristics of the bases are the key drivers of \snp’s empirical performance, yielding dictionaries whose effective power is on par with the intuitive eigenmodes learned by \strainer.

\begin{figure}[t!]
    \centering
    \includegraphics[width=\linewidth]{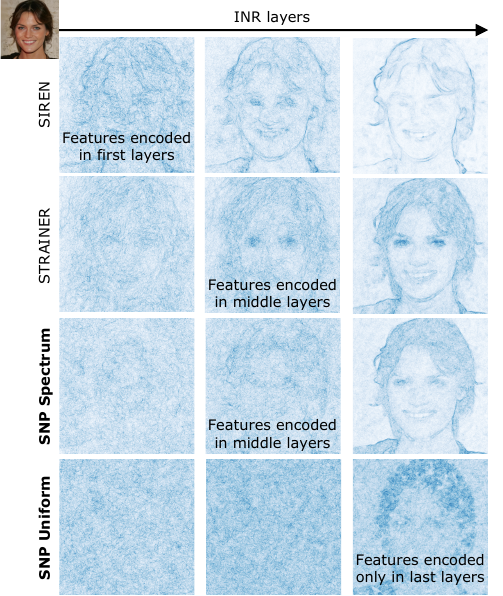}
    \caption{\textbf{Underlying functional local complexity of INR layers.}. Using a measure of local complexity (LC)~\cite{humayun2024deep}, we visualize learned geometry of \siren, \strainer, and \snp INRs fit to a test signal from \celeba. \siren's internal function geometry aligns with the signal morphology. \strainer and \snp: Spectrum exhibit geometry of their pretraining signals in initial layers, while later layers are adapted towards the test signal morphology. Finally, seemingly random geometry is exhibited by initial layers of \snp: Uniform much due to its pretraining on uniform noise.}
    \label{fig:splines}
\end{figure}

\paragraph{Local complexity analysis.} Using spline theory~\cite{balestriero18b, humayun2023splinecam}, we can study properties of a network's underlying function geometry manifold. We in particular focus on the Local Complexity (LC) measure proposed by Humayun et.al.~\cite{humayun2024deep}, which measures the density of input space partitions, to study training dynamics of learned \snp INRs pretrained on Uniform and Spectrum noises.
Fig.~\ref{fig:splines} visualizes the approximated input space partitions as an \snp INR fits to a given signal. High LC (darker) regions indicate high local non-linearity in the INR internal geometry, and generally correlate with high frequency signal details,
while lighter regions denote regions of low LC and correlate with flat or smooth signal regions. 

A trained INR's initial layers tend to reflect the features of its training data. A \siren INR trained on a \celeba image encodes facial features with LC strongly correlated to that signal. In contrast, a \strainer INR trained on a pool of face images exhibits LC correlated to the average face, similar to the eigenvectors in Fig.~\ref{fig:ntk_eigvecs}. Since \snp: Spectrum images are relatively smooth, the LCs of resulting initial INR layers are also smooth with no strong correlation to image edges. Because of this, \snp: Spectrum must compose some finer facial features in the middle layers. \snp: Uniform exhibits uncorrelated and higher LC (darker regions) in initial INR layers, unlike \snp: Spectrum, \strainer, and \siren, indicating that it encodes the input space (coordinates) into more complex and random features. Deeper layers also show high LC, but with strong correlation to signal morphology.

\begin{figure}[t]
    \centering
    \includegraphics[width=\linewidth]{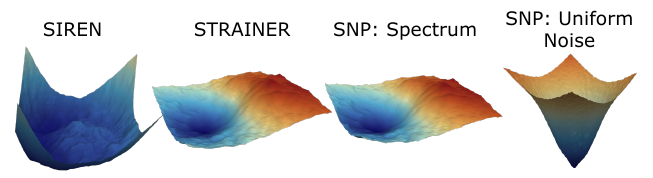}
    \caption{\textbf{Visualizing loss landscapes of \snp INRs.} SNP: Uniform yields
a smooth landscape with a strong minima, justifying its impressive fitting results. STRAINER and SNP: Spectrum exhibit strikingly similar loss surfaces.}
    \label{fig:loss_landscape}
\end{figure}

\paragraph{Loss Landscape Visualization.}
We further visualize loss landscapes~\cite{visualloss} in Figure~\ref{fig:loss_landscape} to understand convergence behavior of \snp INRs.
\snp: Uniform yields a smooth loss landscape with a distinctive minimum, resulting in its rapid convergence to a high-quality signal representation. Interestingly, \snp: Spectrum and \strainer yield strikingly similar landscapes, largely due to the $1/|f^\alpha|$-encoded prior, while \siren yields a rugged landscape without a clear minimum. 
These findings are consistent with image fitting results in Table~\ref{tab:metrics_jumbo}, and denoising results in Table~\ref{tab:denoising_quantitative}, which highlight \snp: Uniform's superior performance for signal fitting and \snp: Spectrum's comparable performance to \strainer for both signal fitting and denoising.

\section{Discussion and Conclusion}
\label{sec:discussion}
Results reported in Table~\ref{tab:metrics_jumbo} and Table~\ref{tab:video_results} first show that pretraining INRs on noise, whether structured or unstructured, yields an extremely powerful initialization for image fitting. \snp{}s with unstructured noise (Uniform or Gaussian) in particular completely outperform all other methods in image fitting, achieving $\approx80$dB for images fairly rapidly. \snp with spectral noise also performs well at image fitting, achieving comparable results to STRAINER. Given that purely unstructured noise is best for image fitting, it is reasonable to assume that spectral noise performs better than Dead Leaves. The trends for video fitting are not as drastic, with \snp with spectral noise performing best, followed closely by \snp: Uniform. 

Fig.~\ref{fig:ntk_curves} demonstrates that signal convergence for \snp: Uniform INRs is mostly explained by the top NTK eigenmodes. This finding empirically explains the fast convergence of \snp: Uniform, further supported by loss landscapes shown in Fig.~\ref{fig:loss_landscape}. From the analysis of learned geometries of INRs (Fig.~\ref{fig:splines}), we further see that the initial layers of \snp: Uniform exhibit highly nonlinear and complex partitions which essentially subdivide the input space into a pseudo-random encoding. We see parallels of this approach with hash-encoding implicit representation frameworks such as Instant-NGP~\cite{ngp} and Diner~\cite{xie2023diner}, which attain impressive fitting capacity by mapping coordinates to random hash codes. With some engineering improvements, it may be possible to mimic the impressive capacity and speed of Instant-NGP with MLPs initialized in the \snp: Uniform style.

\begin{figure}
    \centering
    \includegraphics[width=\linewidth]{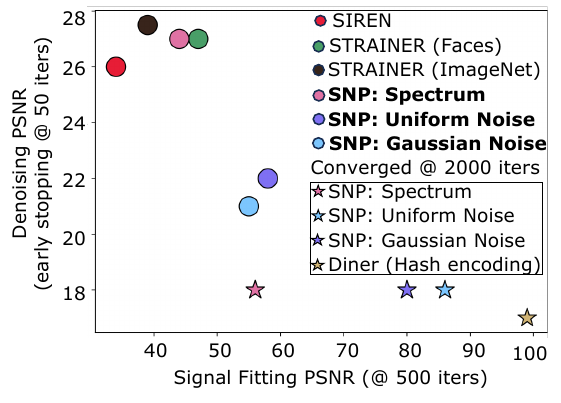}
    \caption{\textbf{Tradeoff in signal representation and denoising ability of \snp INRs on \celeba images.} At early stages of signal fitting ($T=500$) and denoising ($T=50$), \snp: Spectrum and \strainer perform well. In contrast, \snp: Uniform and Gaussian excel at signal fitting, but are poor denoisers. SIREN is a good denoiser but a poor signal fitter. Hence, there is a tradeoff to consider when choosing an appropriate initialization method for a task.}
    \label{fig:inv_tradeoff}
\end{figure}

However, considering signal fitting alone does not tell the whole story. Results in Table~\ref{tab:denoising_quantitative} show that \snp with unstructured noise performs worst of all methods at image denoising. This may be explained by the fact that randomness in their functional embeddings (as shown in Fig.~\ref{fig:splines}) is not conducive to encoding structural priors needed for solving inverse problems. Hence, there is a clear tradeoff that must be considered when choosing INR parameter initialization methods, balancing both functional capacity and strength of its deep priors (see Fig.~\ref{fig:inv_tradeoff}). Fig.~\ref{fig:inv_tradeoff} displays this tradeoff for \celeba images. Of all noise types, \snp with spectral noise offers the best overall performance on both tasks. 

\emph{A main recommendation of this study is that \snp with spectral noise of $1/|f^\alpha|$ frequency structure is an excellent and cheap way to initialize INRs for natural signal fitting and reconstruction tasks. This approach is a particularly attractive alternative to current data-driven initialization methods that require prior real signals.}

A key limitation of this study is that we considered only the sinusoidal activation function and a fixed number of layers. Several prior works have shown that alternative activation functions can result in faster training times and could be better suited for inverse problems. A promising future study lies in understanding how the choice of activation function and network depth interact with optimal choice of training noise structure for INR parameter pretraining.

\section*{Acknowledgments}
This research is supported by NIH DeepDOF R01DE032051 and the UCR Winston Chung Global Energy Center (WCGEC) seed grant. Supported by the Intelligence Advanced Research Projects Activity (IARPA) via Department of Interior/ Interior Business Center (DOI/IBC) contract number 140D0423C0076. The U.S. Government is authorized to reproduce and distribute reprints for Governmental purposes notwithstanding any copyright annotation thereon. Disclaimer: The views and conclusions contained herein are those of the authors and should not be interpreted as necessarily representing the official policies or endorsements, either expressed or implied, of IARPA, DOI/IBC, or the U.S. Government.

\FloatBarrier
{
    \small
    \bibliographystyle{ieeenat_fullname}
    \bibliography{main}
}

\clearpage

\twocolumn[{%
\renewcommand\twocolumn[1][]{#1}%
\maketitlesupplementary
\begin{center}
    \centering
    \captionsetup{type=figure}
    \includegraphics[]{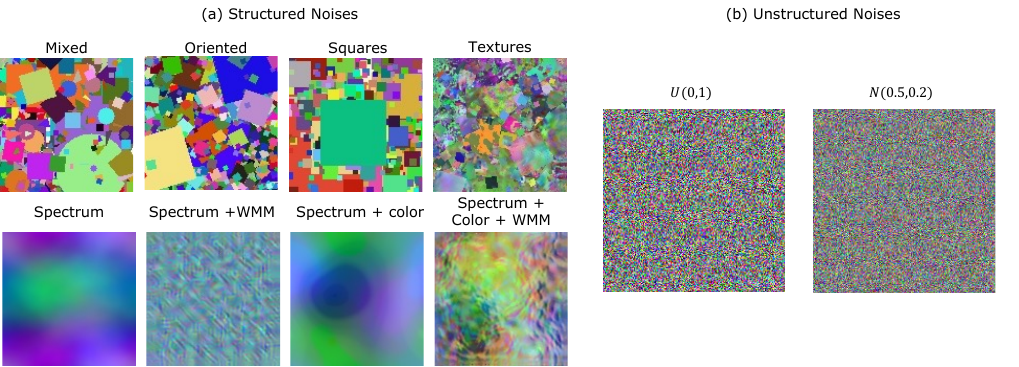}
    \caption{\textbf{Example images used \snp training dataset}(a) Structured noise images, from dead leaves and statistical image generation models , from Looking at Noise~\cite{baradad2021learning}. (b) Unstructured noise images, Uniform and gaussian noise.}
    \label{fig:supl_viewdataset}

\end{center}%
}]

\section{Structured and Unstructured noises used for pretraining}
\label{supl:supl_viewdataset}

Figure~\ref{fig:supl_viewdataset} showcase example images used for \snp pretraining for both structured (from Baradad et.al~\cite{baradad2021learning}) and unstructured noises (uniform and Gaussian). For more details on how structured noise images are generated, we refer the reader to Baradad et.al~\cite{baradad2021learning}.

\section{\snpB pretrain and fitting procedures} \label{supl:train_fit_derivation}
We first provide the reader with an overview of conventional INRs followed by detailed training and fitting procedures for \snp INRs.

\textbf{Conventional INRs.} Typically an INR, denoted as $f_\theta(\cdot)$ with parameters $\theta$ learns a mapping between $d$ dimensional input coordinates, $\mathbf{x} \in \mathcal{R}^d$, and $D$ dimensional signal, $\mathbf{y} \in \mathcal{R}^D$. INR parameters $\theta$ are iteratively optimized by minimizing an objective function, such as mean squared error, to yield the optimal learned signal representation $f_{\theta^*}$.
\begin{align}
\label{eq:inr_basics}
    \theta^* = \arg\min_{\theta} \mid\mid f_\theta(\mathbf{x}) - \mathbf{y}(\mathbf{x})\mid\mid^2_2
\end{align}

\textbf{\snpB INRs.} \snp follows the design of the \strainer~\cite{vyas_strainer} INR framework. As described in \strainer~\cite{vyas_strainer}, at pretraining time, \snp shares initial $L-1$ \emph{encoder} layers while jointly fitting $N$ signals, each with their respective signal specific decoder. At test-time, \snp retains the pretrained encoder weights as an initialization along with  randomly initialized decoder and proceeds to fit an unseen signal. 

We define the initial encoder layers of \snp as $f_\theta(\cdot)$. During pretraining stage, for each train signal $\mathbf{y}^i$, we define signal specific decoders denoted by $g_\psi^i(\cdot)$. We define \snp INR $h_{\Theta}(\cdot)$ below in Eq.~\ref{eq:supl_snp}

\begin{align}
    \label{eq:supl_snp}
    h_\Theta(\mathbf{x}) = g^{i}_\psi \circ f_\theta(\mathbf{x}) 
\end{align}

where $\Theta$ is the collective set of encoder weights $\theta$ and decoder weights $\psi^1 \cdots \psi^N $. 

During pretraining, we fit $N$ signals of the same noise-type jointly to \snp for $T$ steps, following Eq~\ref{eq:supl_pretrain_snp}, to obtain learned encoder initialization $\theta^*$

\begin{align}
    \Theta^* = \arg\min_{\Theta} \sum_{i=1}^{i=N} \mid\mid g_\psi^i \circ f_\theta(\mathbf{x}) - \mathbf{y^i}(\mathbf{x}) \mid\mid^2_2
    \label{eq:supl_pretrain_snp}
\end{align}

where $\Theta^*  = [\theta^*, \psi^{1*}, \cdots ,\psi^{N*}]$

At test-time, \snp is initialized learned encoder weights $f_{\theta=\theta^*}$ along with a randomly initialized  decoder $g_{\psi}$. Finally, we fit the test-signal for $T$ steps to the pretrained \snp INR as shown in Eq.~\ref{eq:supl_testime_fit_snp}. 

\begin{align}
    [\theta^*, \psi^*] \leftarrow \arg\min_{\theta, \psi} \mid\mid g_\psi \circ f_{\theta=\theta^*} (\mathbf{x}) - \mathbf{y}(\mathbf{x}) \mid\mid^2_2
    \label{eq:supl_testime_fit_snp}
\end{align}

\section{Training specifications} \label{supl:training_specs}
\subsection{Image and Video fitting}

\textbf{Image fitting.} We describe the pretraining and fitting procedure for \snp INRs. For pretraining, we follow the process outlined in \strainer~\cite{vyas_strainer} and use $N=10$ images  (which has previously shown success in learning high quality initialization in \strainer) for each type of noise (illustrated in Fig.~\ref{fig:supl_viewdataset}). At test-time, we retain the pretrained features as initialization of the \emph{encoder} or first $L-1$ layers and randomly initialize the last \emph{decoder} layer. We fit each test image for $2000$ iterations using Adam optimizer with learning rate of $10^{-4}$. All hidden layers are of width $256$. 

\textbf{Video fitting.} We adopt ResField as our video-fitting backbone to apply the \snp initialization. ResField’s parameters are decomposed into temporally shared weight and per-frame low-rank residuals. We transfer the weights learned during \snp noise pre-training to the shared branch, while initializing the residual factors to zero \cite{resfield,lora}, i.e., $A\sim\mathcal{N}(0,\sigma^{2}), \, B = 0,\, \Delta W = BA = 0 \text{ at the first iteration}$. We employ an MLP with four hidden layers of width 256 for the shared weights and rank-10 residual weights. 
Each video is optimized for \(10^{5}\) iterations using the Adam optimizer 
(learning rate \(5 \times 10^{-4}\)) and a cosine-annealing learning-rate schedule. We perform pre-training on noise images rather than noise videos, and then successfully transfer the resulting model weights from the image domain to the video domain. This strategy is highly practical, as it substantially reduces both pre-training time and the complexity of pretraining data generation.

\textbf{Image and Video Denoising.} For image and video denoising, we simulate a Poisson process to add synthetic noise of photon count of 30, resulting in a read-out noise of 2 dB to each image/video frame. At test-time we initialize the \snp encoder using the corresponding weights from fitting tasks.

We use the Alpine INR framework~\cite{vyasalpine} for pretraining and test-time signal fitting for \snp INRs.

\section{More results on image and video fitting} \label{more_results_image_fitting}
\begin{figure}[h]
    \centering
    \includegraphics[]{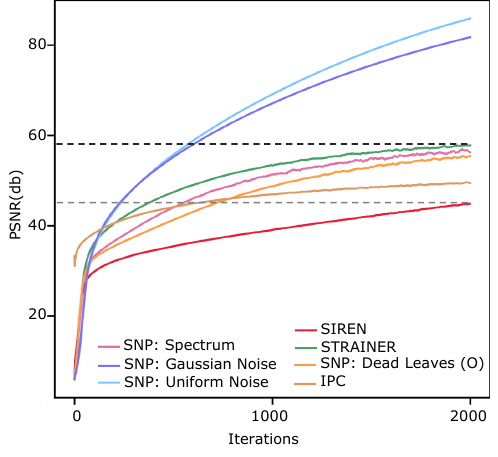}
    \caption{\textbf{Signal fitting results on \afhq images.} \snp initialized using various noises demonstrate strong signal representation ability and convergence compared to widely known baseline methods such as \siren~\cite{sitzmann2020implicit}, \ipc~\cite{ipc}, and \strainer~\cite{vyas_strainer}. Interestingly, \snp trained with Gaussian and Uniform noises performed significantly better than all other approaches. converging rapidly to perfect fitting.}
    \label{fig:supl_afhq_curves}
\end{figure}

\begin{figure}[h]
    \centering
    \includegraphics[]{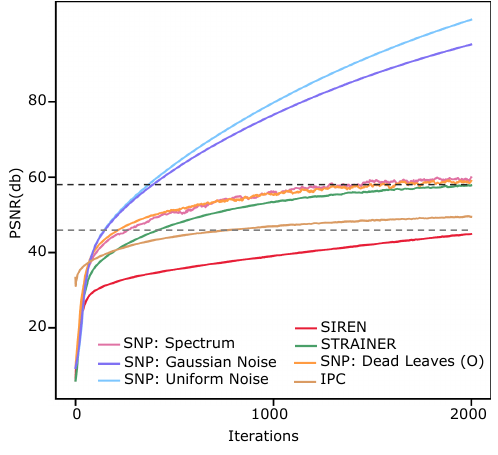}
      \caption{\textbf{Signal fitting results on \mri images.} \snp initialized using various noises demonstrate strong signal representation ability and convergence compared to widely known baseline methods such as \siren~\cite{sitzmann2020implicit}, \ipc~\cite{ipc}, and \strainer~\cite{vyas_strainer}. Interestingly, \snp trained with Gaussian and Uniform noises performed significantly better than all other approaches. converging rapidly to perfect fitting.}
    \label{fig:supl_mri_curves}
\end{figure}

Table~\ref{tab:metrics_jumbo} (in main text) presented a brief snapshot of our experimental evaluation. We provide a more detailed analysis in Tables~\ref{tab:supl_celeba_metrics},~\ref{tab:supl_afhq_metrics}, and Table~\ref{tab:supl_mri_metrics} and report mean values of PSNR, SSIM, and LPIPS~\cite{lpips} and 1 std dev. for randomly initialized \siren, data-driven baselines such as \strainer, and \snp variants for image fitting on \celeba,\afhq, and \mri datasets. We observe that \snp pretrained on uniform and Gaussian noise strikingly, achieves a PSNR of $80db+$, outperforming all other methods. On structured noise initializations, \snp: Spectrum performs best, almost matching \strainer. Figures~\ref{fig:supl_mri_curves} and ~\ref{fig:supl_afhq_curves} also show image fitting trajectories for \afhq and \mri datasets respectively and we observe that \snp pretrained on all noise types to quickly converge to high quality signal representations. Notably, \snp: Uniform achieves \siren level performance $20\times$ faster and \strainer level performance $4\times$ faster when evaluated on \celeba( see Fig.~\ref{fig:results_celeba_psnr_plot} in main text) and on \afhq and \mri datasets as shown in Fig.~\ref{fig:supl_afhq_curves} and Fig.~\ref{fig:supl_mri_curves} in attached supplement.

\begin{table}[H]
\centering
\caption{\textbf{Image fitting results on \celeba.} We present mean PSNR(in db), SSIM and LPIPS with 1 std.dev. for all test methods at $T=2000$ iterations, from vanilla randomly initialized (\siren), to data driven methods such as Metalearned~\cite{tancik2021learned}, \strainer~\cite{vyas_strainer}, and \snp variants trained on structured and unstructured noises.}
\label{tab:supl_celeba_metrics}

\small
\setlength{\tabcolsep}{2pt}
\scalebox{0.7}{
\begin{tabular}{lc|c|c}
\toprule
\multirow{2}{*}{Method} &
\multicolumn{3}{c}{CelebA} \\
\cmidrule(lr){2-4}
& PSNR$\uparrow$ & SSIM$\uparrow$ & LPIPS$\downarrow$ \\
\midrule

\siren~\cite{sitzmann2020implicit} &
$44.9 \pm 2.13$ & $0.991 \pm 0.007$ & $0.0026 \pm 0.002$ \\

\strainer~\cite{vyas_strainer} &
$57.8 \pm 3.46$ & $0.999 \pm 0.001$ & $0.00002 \pm 0.07$ \\

Meta-learned~\cite{tancik2021learned} &
$53.1 \pm 3.36$ & $0.994 \pm 0.053$ & $0.0046 \pm 6.2\text{e-5}$ \\

\midrule
\snp: Dead leaves oriented &
$55 \pm 2.8$ & $0.999 \pm 0.0006$ & $0.00096 \pm 0.001$ \\

\snp: Dead leaves texture &
$55 \pm 2.9$ & $0.999 \pm 0.0006$ & $0.0009 \pm 0.001$ \\

\snp: Dead leaves mixed &
$54.6 \pm 3.0$ & $0.999 \pm 0.0007$ & $0.0011 \pm 0.001$ \\

\snp: Dead leaves squares &
$54.7 \pm 2.7$ & $0.999 \pm 0.0007$ & $0.0012 \pm 0.001$ \\

\midrule
\snp: Spectrum &
$56.4 \pm 3.1$ & $0.999 \pm 0.0006$ & $0.0005 \pm 0.001$ \\

\snp: Spectrum + Color &
$51 \pm 2.5$ & $0.998 \pm 0.0011$ & $0.0039 \pm 0.003$ \\

\snp: Spectrum + WMM &
$54.4 \pm 3.1$ & $0.999 \pm 0.0007$ & $0.0011 \pm 0.001$ \\

\snp: Spectrum + WMM + Color &
$55.2 \pm 3.4$ & $0.999 \pm 0.0007$ & $0.0008 \pm 0.0017$ \\

\midrule

\snp: Gaussian  &
$80 \pm 11.2$ & $0.999 \pm 4.20\text{e-5}$ & $5.50\text{e-6} \pm 1.5\text{e-5}$ \\

\textbf{\snpB: Uniform } &
$\mathbf{85.7 \pm 12.6}$ & $\mathbf{0.999 \pm 2.90\text{e-5}}$ & $\mathbf{3.40\text{e-6} \pm 1.10\text{e-5}}$ \\

\bottomrule
\end{tabular}
}
\end{table}

\begin{figure}[H]
   \centering
   \includegraphics[width=\linewidth]{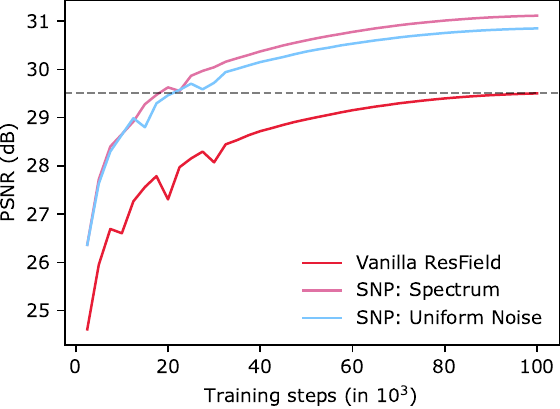}
   \caption{\textbf{Video fitting results on the all video in our dataset.}. \snp initialization accelerates the training of ResField by approximately a factor of five compared to the vanilla model across all videos.}
   \label{fig:vid_fitting_curves}
\end{figure}

\begin{figure*}[!t]
    \includegraphics[width=\linewidth]{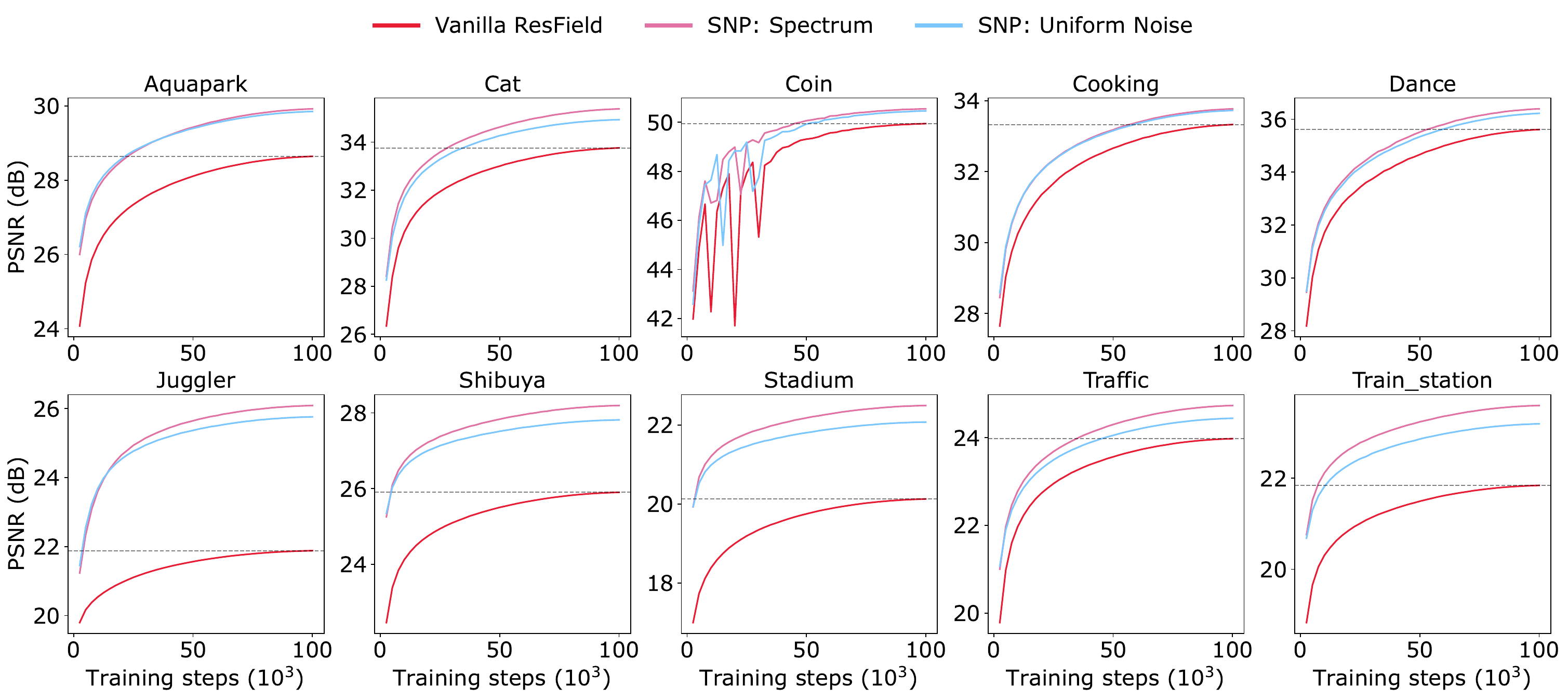}
    \caption{The \snp pretrained model provides a superior initialization, resulting in substantial starting gains and a consistently faster convergence rate during test-time fine-tuning across all videos.}
    \label{fig:individual_fitting}
\end{figure*}

Table~\ref{tab:supl_video_detailed_results} summarizes video fitting results. To illustrate convergence efficiency, Fig~\ref{fig:vid_fitting_curves} plots the learning curve averaged across the dataset; notably, our \snp-initialized model attains the baseline's 100k-iteration performance in just 20k iterations. The corresponding learning curves for individual videos are detailed in \cref{fig:individual_fitting}.

\begin{table*}
\centering
\caption{PSNR (dB) comparison across scenes for video representation and video denoising.}
\label{tab:supl_video_detailed_results}
\resizebox{\textwidth}{!}{%
\begin{tabular}{llrrrrrrrrrrr}
\hline
 Task & Method & Cat & Coin & Cooking & Dance & Aquapark & Stadium & Train & Shibuya & Juggler & Traffic & All\\
\hline
\multirow{3}{*}{Fitting}
& Vanilla Resfield            & 34.1 & 50.0 & 33.5 & 35.5 & 28.6 & 20.1 & 21.8 & 25.9 & 21.9 & 24.0 & 29.5\\
& ResField+\snp (Uniform) & 35.1 & 50.5 & 33.9 & 36.1 & 29.8 & 22.1 & 23.2 & 27.8 & 25.7 & 24.4 & 30.9\\
& ResField+\snp(Spectrum)            & 35.6 & 50.5 & 33.9 & 36.3 & 29.9 & 22.5 & 23.6 & 28.2 & 26.1 & 24.7 & 31.1\\
\hline
\multirow{3}{*}{Denoising}
& Vanilla Resfield            & 32.2 & 37.5 & 31.6 & 29.7 & 27.1 & 19.9 & 21.6 & 25.4 & 21.5 & 23.5 & 27.0\\
& ResField+\snp (Uniform) & 32.8 & 34.3 & 31.6 & 28.5 & 27.9 & 21.8 & 22.8 & 27.2 & 24.8 & 23.9 & 27.6\\
& ResField+\snp (Spectrum)           & 33.3 & 35.4 & 31.8 & 29.1 & 28.0 & 22.2 & 23.2 & 27.5 & 25.1 & 24.1 & 28.0\\
\hline
\end{tabular}%
}
\end{table*}

\begin{table}
\centering
\caption{\textbf{Image fitting results on \afhq.} We present mean PSNR(in db), SSIM and LPIPS with 1 std.dev. for all test methods at $T=2000$ iterations, from vanilla randomly initialized (\siren), to data driven methods such as Metalearned~\cite{tancik2021learned}, \strainer~\cite{vyas_strainer}, and \snp variants trained on structured and unstructured noises.}
\label{tab:supl_afhq_metrics}

\small
\setlength{\tabcolsep}{2pt}
\scalebox{0.7}{
\begin{tabular}{lc|c|c} 
\toprule
\multirow{2}{*}{Method} &
\multicolumn{3}{c}{AFHQ} \\
\cmidrule(lr){2-4}
& PSNR$\uparrow$ & SSIM$\uparrow$ & LPIPS$\downarrow$ \\
\midrule

\siren~\cite{sitzmann2020implicit} &
$45.1 \pm 3.13$ & $0.991 \pm 0.005$ & $0.0025 \pm 0.004$ \\

\strainer~\cite{vyas_strainer} &
$58 \pm 3.75$ & $0.999 \pm 0.001$ & $0.00003 \pm 0.06$ \\

Meta-learned~\cite{tancik2021learned} &
$53.3 \pm 2.52$ & $0.996 \pm 0.044$ & $0.003 \pm 8.0\text{e-5}$ \\

\midrule
\snp: Dead leaves oriented &
$55.4 \pm 2.93$ & $0.999 \pm 0.0006$ & $0.0008 \pm 0.0011$ \\

\snp: Dead leaves texture &
$55.3 \pm 3.5$ & $0.999 \pm 0.0007$ & $0.0007 \pm 0.001$ \\

\snp: Dead leaves mixed &
$55 \pm 3.1$ & $0.999 \pm 0.0007$ & $0.0008 \pm 0.0011$ \\

\snp: Dead leaves squares &
$55 \pm 3.3$ & $0.999 \pm 0.0008$ & $0.0009 \pm 0.0015$ \\

\midrule
\snp: Spectrum &
$56.2 \pm 3.3$ & $0.999 \pm 0.00045$ & $0.0004 \pm 0.0013$ \\

\snp: Spectrum + Color &
$51.1 \pm 2.9$ & $0.998 \pm 0.0014$ & $0.003 \pm 0.0035$ \\

\snp: Spectrum + WMM &
$54.7 \pm 3.5$ & $0.999 \pm 0.00088$ & $0.0009 \pm 0.0015$ \\

\snp: Spectrum + WMM + Color &
$55.9 \pm 3.3$ & $0.999 \pm 0.0006$ & $0.0005 \pm 0.00087$ \\

\midrule
\snp: Gaussian &
$81.8 \pm 13.1$ & $0.999 \pm 4.70\text{e-5}$ & $3.96\text{e-6} \pm 1.06\text{e-5}$ \\

\textbf{\snpB: Uniform} &
$\mathbf{85.91 \pm 14.4}$ & $\mathbf{0.999 \pm 3.30\text{e-5}}$ & $\mathbf{2.60\text{e-6} \pm 7.70\text{e-6}}$ \\

\bottomrule
\end{tabular}
}
\end{table}

\begin{table}
\centering
\caption{\textbf{Image fitting results on \mri.} We present mean PSNR(in db), SSIM and LPIPS with 1 std.dev. for all test methods at $T=2000$ iterations, from vanilla randomly initialized (\siren), to data driven methods such as Metalearned~\cite{tancik2021learned}, \strainer~\cite{vyas_strainer}, and \snp variants trained on structured and unstructured noises.}
\label{tab:supl_mri_metrics}

\small

\setlength{\tabcolsep}{2pt}
\scalebox{0.7}{
\begin{tabular}{lc|c|c}
\toprule
\multirow{2}{*}{Method} &
\multicolumn{3}{c}{OASIS-MRI} \\
\cmidrule(lr){2-4}
& PSNR$\uparrow$ & SSIM$\uparrow$ & LPIPS$\downarrow$ \\
\midrule

\siren~\cite{sitzmann2020implicit} &
$53.03 \pm 1.72$ & $0.999 \pm 0.0002$ & $0.0009 \pm 0.0002$ \\

\strainer~\cite{vyas_strainer} &
$62.8 \pm 3.17$ & $0.999 \pm 0.0003$ & $0.00001 \pm 2.0\text{e-5}$ \\

Meta-learned~\cite{tancik2021learned} &
$67 \pm 2.27$ & $0.999 \pm 1.0\text{e-5}$ & $0.000005 \pm 2.5\text{e-6}$ \\

\midrule
\snp: Dead leaves oriented &
$59 \pm 3.5$ & $0.999 \pm 0.0001$ & $0.0003 \pm 0.0003$ \\

\snp: Dead leaves texture &
$58.9 \pm 3.38$ & $0.999 \pm 0.0002$ & $0.0003 \pm 0.0003$ \\

\snp: Dead leaves mixed &
$59.5 \pm 2.76$ & $0.999 \pm 9.7\text{e-5}$ & $0.0002 \pm 0.00019$ \\

\snp: Dead leaves squares &
$59.1 \pm 3.3$ & $0.999 \pm 0.0001$ & $0.0002 \pm 0.00026$ \\

\midrule
\snp: Spectrum &
$60 \pm 3.9$ & $0.999 \pm 0.0002$ & $0.00017 \pm 0.0004$ \\

\snp: Spectrum + Color &
$56.2 \pm 2.69$ & $0.999 \pm 0.0002$ & $0.00098 \pm 0.0003$ \\

\snp: Spectrum + WMM &
$59 \pm 3.3$ & $0.999 \pm 1.0\text{e-4}$ & $0.0002 \pm 0.0002$ \\

\snp: Spectrum + WMM + Color &
$59.8 \pm 3.3$ & $0.999 \pm 9.5\text{e-5}$ & $0.00018 \pm 0.0002$ \\

\midrule

\snp: Gaussian  &
$95.2 \pm 5.2$ & $0.999 \pm 1.40\text{e-7}$ & $9.7\text{e-8} \pm 4.14\text{e-8}$ \\

\textbf{\snpB: Uniform } &
$\mathbf{101.8 \pm 6.03}$ & $\mathbf{0.999 \pm 6.30\text{e-8}}$ & $\mathbf{6.90\text{e-8} \pm 1.9\text{e-8}}$\\

\bottomrule
\end{tabular}
}
\end{table}

\section{ Number of images used for learning \snp encoder } 
\label{supl:ablation_num_images}

\begin{figure}[H]
   \centering
   \includegraphics[width=\linewidth, height=2.5in]{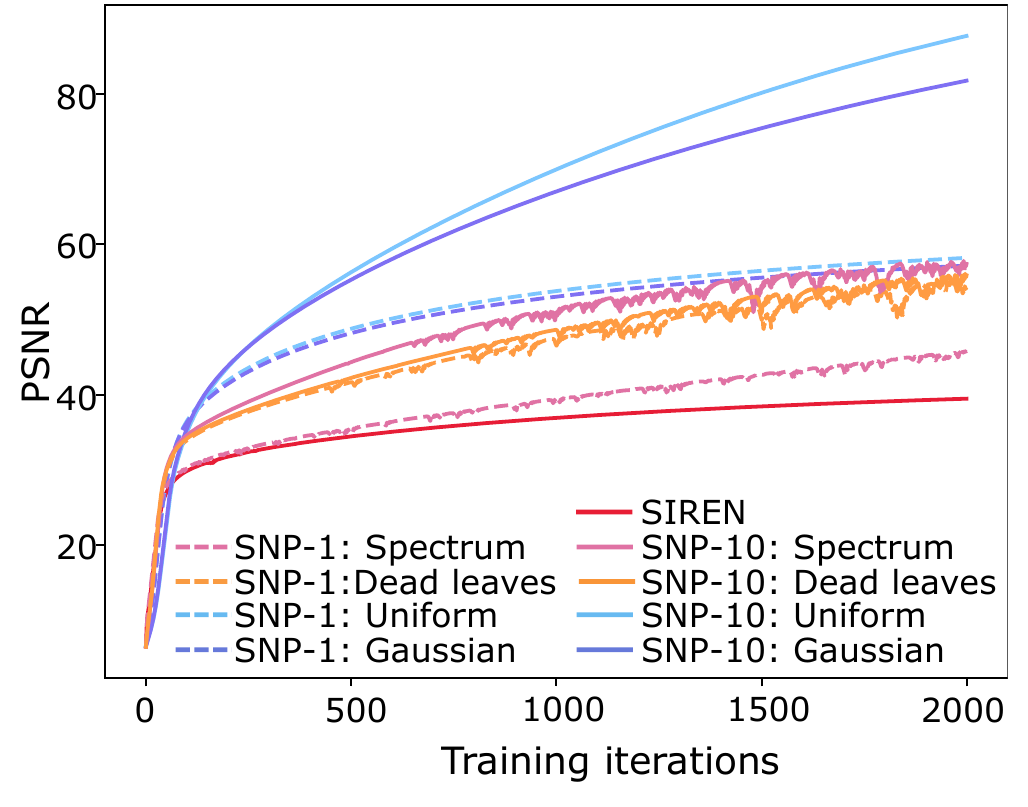}
   \caption{\textbf{Fitting performance v/s number of pretraining signals.} We show that initialization learned by \snp pretrained on multiple pretraining images, such as $N=10$, learns a richer and powerful initialization achieving significantly better reconstruction quality than \snp pretrained on $N=1$ signals. (\snp-$k$ denotes INRs pretrained on $k$ images).}
   \label{fig:supl_snp1_100}
\end{figure}

We study how the signal fitting performance changes based on number of training images used for pretraining \snp. We run a simple evaluation of \snp pretrained using $N=1$ images (\snp-1) and $N=10$ images (\snp-10) and fit to $10$ samples from the \celeba dataset and plot reconstruction PSNR as a function of training iterations, shown in Fig~\ref{fig:supl_snp1_100}. We observe that \snp pretrained on even $1$ unstructured noise image (uniform and Gaussian) achieves comparable performance to \snp-10 pretrained on spectrum and dead leaves images. However, \snp-10 pretrained on $10$ unstructured noise images perform significantly better in comparison, out performing \snp-1 (Uniform) by $\approx 40db+$. 

\section{Extending \snp to other activations (FINER)} \label{sec:supl_finer}
We extend \snp to FINER~\cite{liu2023finer} and demonstrate its signal fitting performance on \celeba dataset. We swap our Sine activations from \snp with the flexible FINER activations proposed by Liu et.al~\cite{liu2023finer}. We pretrain \snp(FINER) on Spectrum and Uniform noise: specifically, we initialize the first-layer biases from $\mathcal{U}\left(-1/\sqrt{2}, 1/\sqrt{2}\right)$ and employ 10 noise images, otherwise adhering to our original hyperparameter setup. Table~\ref{tab:supl_finer_metrics_celeba} reports a baseline FINER INR along with \snp(FINER) pretrained on spectrum and Uniform noises. We find \snp (FINER) pretrained on Uniform noise achieves a staggering  PSNR of $100db+$ on \celeba faces. \snp trained on both, Uniform and Spectrum noises, outperform FINER~\cite{liu2023finer} further demonstrating the effectiveness  of noise pretraining and \snp's ability to readily  extend to other activation functions.

\begin{table}[H]
    \centering
    \caption{\textbf{\snp readily extends to FINER activations~\cite{liu2023finer}.} We extend \snp framework to FINER~\cite{liu2023finer} activations and evaluate image fitting performance on \celeba.}
    \label{tab:supl_finer_metrics_celeba}
    \resizebox{\linewidth}{!}{
    \begin{tabular}{|c|c|c|c|}
    \hline
    Method &  PSNR & SSIM & LPIPS \\
    \hline
         FINER~\cite{liu2023finer} & $51.2 \pm 3.1$ & $0.998 \pm 0.0012$ & $0.001 \pm 0.002$ \\
         \snp Finer: Spectrum & $52.3 \pm 2.4$ & $0.998 \pm 0.001$ & $0.003 \pm 0.002$\\
         \snpB \textbf{Finer: Uniform} & $\mathbf{117.2 \pm 14}$ & $\mathbf{0.999 \pm 2.1\mathrm{E}{-6}}$ & 
         $\mathbf{1.8\mathrm{E}{-7} \pm 2.1\mathrm{E}{-6}}$ \\
     \hline
    \end{tabular}}
\end{table}

\section{Additional baselines and deeper networks} \label{sec:supl_addnexplayers}

We pretrain \strainer using $1000$ images ($100\times$ than \snp and \strainer), each randomly selected from a different class to ensure spectral coverage,  for $10^5$ epochs using the best model for image fitting. Table~\ref{tab:rb_result_table} shows \snp: Spectrum outperforming \strainer: ImageNet, indicating the crucial role of $1/f$ prior directly encoded through spectrum noise.  Table~\ref{tab:rb_result_table} shows \snp also generalizes to deeper 10-layer INRs.

\begin{table}[h]
\centering
\caption{\textbf{Fitting results} on additional baselines and deeper INRs.}
\label{tab:rb_result_table}
\setlength{\tabcolsep}{1.8pt}
\scalebox{0.8}{
\begin{tabular}{lccc}
\hline
\textbf{Model} & \textbf{CelebA-HQ} & \textbf{AFHQ} & \textbf{OASIS-MRI} \\ \hline
    \siren & 44.9 & 45.1 & 53.03 \\
    \strainer: ImageNet & 50.87 & 44.11 & 53.37 \\
    \strainer (dataset-specific) & 57.8 & 58.0 & 62.80 \\
    \textbf{\snp: Spectrum} & 56.4 $\uparrow$ & 56.2 $\uparrow$ & 60.0 $\uparrow$\\
    \textbf{\snp: Uniform} & 85.7 $\uparrow$& 79.9 $\uparrow$& 79.3 $\uparrow$\\ 
\hline
    \siren (10 layers) & 51.3 & 52.6  & 57.3 \\
    \textbf{\snp: Spectrum (10 layers)} & 64.1 $\uparrow$ & 64.7 $\uparrow$ & 66.2 $\uparrow$ \\
    \textbf{\snp: Uniform (10 layers)} & 100+ $\uparrow$ & 100+ $\uparrow$ & 100+ $\uparrow$ \\
\hline
\end{tabular}
}
\end{table}

\section{Visualizing \snpB initial and learned geometry} \label{supl:vis_snp_geometry}

\begin{figure*}[t!]
    \includegraphics[]{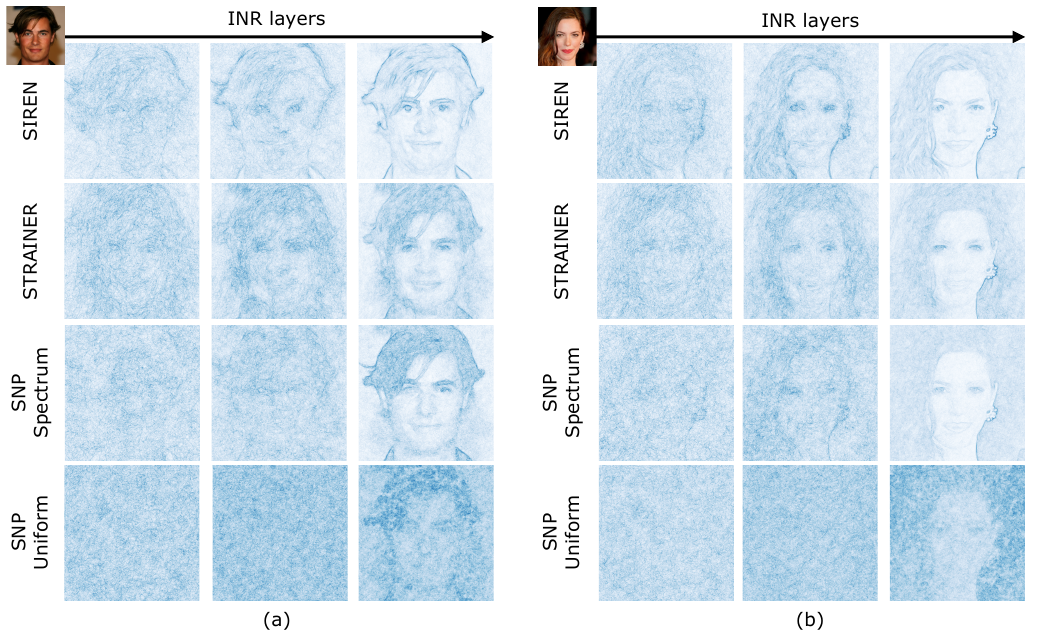}
    \caption{\textbf{More examples on local complexity of INR layers for \snpB and baselines.} Using local complexity measure~\cite{humayun2024deep}, we visualize , in (a) and (b), the local complexity of \siren, \strainer and \snp (uniform and spectrum) fit to images from \celeba . We observe that \siren's internal layers quickly adapt to the underlying signal morphology. \strainer and \snp (spectrum) exhibit internal geometry akin to their pretraining signals. Finally, we observe that \snp Uniform exhibit seemingly random geometry in the initial layers, much due to its pretraining on uniform noise.}
    \label{fig:supl_addn_splines}
\end{figure*}

Following our local complexity (LC)  analysis presented in Sec.~\ref{sec:experiments_results} (of main text), we observe similar trends in the learned geometry of INR layers on multiple samples from the \celeba dataset as shown in Fig.~\ref{fig:supl_addn_splines}. Similar to Fig~\ref{fig:splines} (of main text), we find that initial layers of the INR tend to reflect the features of their respective pretraining data. \siren, which is solely fit to a given signal, exhibits LC strongly aligned with the signal morphology. We find \strainer, which is pretrained on \celeba faces, to exhibit LC more similar to an average of all pretraining images. Furthermore, \snp spectrum , pretrained smooth noise images with $1/f^\alpha$ structure, reveals LC of initial layers to be smooth while later layers are seen to be quickly adapting to the signal structure. Finally, \snp Uniform, showcases a seemingly random LC in initial layers, while the deeper layer reflects extremely high non-linearity compared to other models spatially co-located with the high frequency details present in the signal.

\end{document}